\documentclass{article}
\usepackage{spconf,amsmath,epsfig,subfigure}

\usepackage{color}
\usepackage{url,hyperref}
\usepackage{graphicx}
\usepackage{booktabs}
\usepackage{multirow}
\usepackage[table,xcdraw]{xcolor}
\usepackage{multicol}
\usepackage{rotating}
\usepackage{lscape}

\usepackage{changes}
\definechangesauthor[name={Vlad}, color=orange]{Vlad}
\definechangesauthor[name={Hanhe}, color=blue]{Hanhe}
\definechangesauthor[name={Dietmar}, color=red]{Dietmar} 

\begin{document}\sloppy

\def\x{{\mathbf x}}
\def\L{{\cal L}}

\title{K\MakeLowercase{on}IQ-10k: Towards an ecologically valid and large-scale IQA database}
%
\name{Hanhe Lin*, Vlad Hosu*\thanks{*Hanhe Lin and Vlad Hosu contributed equally.} and Dietmar Saupe}
\address{Department of Computer and Information Science, University of Konstanz, Germany\\
Email: \{hanhe.lin, vlad.hosu, dietmar.saupe\}@uni-konstanz.de}

\maketitle
\begin{abstract}
The main challenge in applying state-of-the-art deep learning methods to predict image quality in-the-wild is the relatively small size of existing quality scored datasets. The reason for the lack of larger datasets is the massive resources required in generating diverse and publishable content. We present a new systematic and scalable approach to create large-scale, authentic and diverse image datasets for Image Quality Assessment (IQA). We show how we built an IQA database, KonIQ-10k\footnote{Database is available at \url{http://database.mmsp-kn.de}.}, consisting of 10,073 images, on which we performed very large scale crowdsourcing experiments in order to obtain reliable quality ratings from  1,467 crowd workers (1.2 million ratings). We argue for its ecological validity by analyzing the  diversity of the dataset, by comparing it to state-of-the-art IQA databases, and by checking the reliability of our user studies.
%
%
%
%
%
%
%
%
\end{abstract}

%
%
%
%
%

\section{Introduction}



Objective Image Quality Assessment (IQA) is important in a broad range of applications, from image compression to display technology and more. To further develop and evaluate objective IQA methods, in particular deep learning methods, large and diverse IQA databases are needed.  ``In research, the ecological validity of a study means that the methods, materials and setting of the study must approximate the real-world that is being examined'' (Wikipedia). The ecological validity of an IQA database refers to the representativeness of the image collection for the wide range of public Internet photos.

Conventionally, creating an IQA database has followed the same typical procedure: collect pristine images and artificially degrade them. Next ask a few volunteers, usually students or naive participants, to assess the quality of the distorted images. The first drawback of the approach is that the diversity of image content is limited since all the distorted images are degraded from a small set of pristine images. Second, the distortions are applied in very limited combinations, whereas ecologically valid distortions are caused by combinations of distortions, also of types that may differ from those in the databases. Last, but not the least, the conventional approach for creating IQA datasets results in small databases, since assessing the quality of a large number of images in a lab setting is too costly.
To address these limitations, we have designed a scalable approach that allowed us to create the largest IQA database to date (images and subjective scores). 
It consists of 10,073 images that were selected from around 10 million YFCC100M \cite{thomee:2016} entries. To ensure the diversity in content and distortions, our sampling algorithm makes use of seven quality indicators (sharpness, colorfulness, ...) and one content indicator (deep features). For each image, 120 reliable quality ratings were obtained by crowdsourcing, performed by 1,467 crowd workers. In comparison to existing IQA databases, ours  contains a vastly larger number of images, with a much broader content diversity and authentic distortions.









\section{Related work}

\begin{table*}[ht]
\footnotesize
\caption{Comparison of existing IQA databases with KonIQ-10k.}
\label{tb:benchmarkdb}
\centering
\resizebox{1.0\textwidth}{!}{
\begin{tabular}{l c r r r c r r c} \hline
& & &\multicolumn{1}{c}{No.\ of}&& No.\ of&\multicolumn{1}{c}{No.\ of}&\multicolumn{1}{c}{Ratings}  \\
Database & Year &Content&distorted images &Distortion type&distortion types &rated images &per image &Environment\\ \hline
IVC \cite{ivcdb}& 2005&10&185& artificial&4&185&15&lab\\
LIVE \cite{sheikh:2006statistical} & 2006 &29&779& artificial & 5&779&23&lab \\
TID2008 \cite{ponomarenko:2009tid2008}&2009&25&1,700& artificial &17&1,700&33&lab \\
CSIQ \cite{larson:2010most} & 2009&30 &866&artificial& 6&866&5$\sim$7&lab \\
TID2013 \cite{ponomarenko:2015image}&2013&25&3,000&artificial&24&3,000&9&lab \\
CID2013 \cite{virtanen2015cid2013}&2013&8&474&authentic&12$\sim$14&480&31&lab \\
LIVE In the Wild \cite{ghadiyaram:2016massive}& 2016&1,169&1,169&authentic&N/A&1,169&175&crowdsourcing \\
Waterloo Exploration \cite{ma2016group}& 2016& 4,744& 94,880 & artificial& 4 & 0&0 & lab \\  \hline
 KonIQ-10k&2017&10,073&10,073&authentic&N/A&10,073&120&crowdsourcing \\
\hline
\end{tabular}
}
\end{table*}

A number of IQA databases have been released in recent years, aiming to help the development and evaluation of  objective IQA methods, see Table~\ref{tb:benchmarkdb}. 

An early conventionally build IQA database, IVC \cite{ivcdb}, was released in 2005. LIVE \cite{sheikh:2006statistical}, TID2008 \cite{ponomarenko:2009tid2008}, and CSIQ \cite{larson:2010most} are the most common databases that researchers use to develop, improve, and evaluate their objective IQA methods. TID2008 was further extended to TID2013 \cite{ponomarenko:2015image} by including seven more distortion types. 
The aforementioned databases, are all small-scale, contain limited content types, and consider few types of artificial distortions. 

Virtanen et al.\ \cite{virtanen2015cid2013} were first to introduce more authentic distortions, created from 480 images of 8 different scenes captured by 79 different cameras. However, the creation method is time-consuming and expensive and thus impractical for large-scale databases. Ghadiyaram et al.\ \cite{ghadiyaram:2016massive} asked a few photographers to capture 1,162 images by a variety of mobile device cameras. Their visual quality was assessed by crowdsourcing experiments. Although this method provides an alternative way to reduce time and cost for IQA subjective study, the database size as well as the content diversity are still relatively low. 

Ma et al.\ \cite{ma2016group} created a database with 4,744 pristine images and 94,880 distorted images 
to validate their proposed mechanism called group MAximum Differentiation (gMAD) competition. 
Their database is meant to provide an alternative evaluation for the performance of IQA models, by means of paired comparisons. Although the Waterloo Exploration database is the largest available in the field, its images are artificially distorted, thus non-authentic, and due to the lack of subjective ratings it cannot be used for developing new IQA methods that rely on them.

In comparison to lab-based studies which are time-consuming and expensive, crowdsourcing has been successfully employed to conduct Quality of Experience (QoE) assessment for images \cite{ghadiyaram:2016massive} and videos \cite{hosu:2017konstanz}.  Although it has been believed that data collected by crowdsourcing is less reliable, 
Redi et al. \cite{siahaan_reliable_2016} verified that crowd workers can generate reliable results under certain  experimental setups. 

\section{Database creation}
\label{sec:databcreate}

\subsection{Overview}

We started from a large public multimedia database, YFCC\-100m \cite{thomee:2016}, from which we randomly selected approximately 10 million (9,974,030) image records. Then, we filtered them down in two stages to the final database of 10,073 images.

\begin{figure}[!t]
\centering
\vspace{-10pt}
\includegraphics[width=0.45\textwidth]{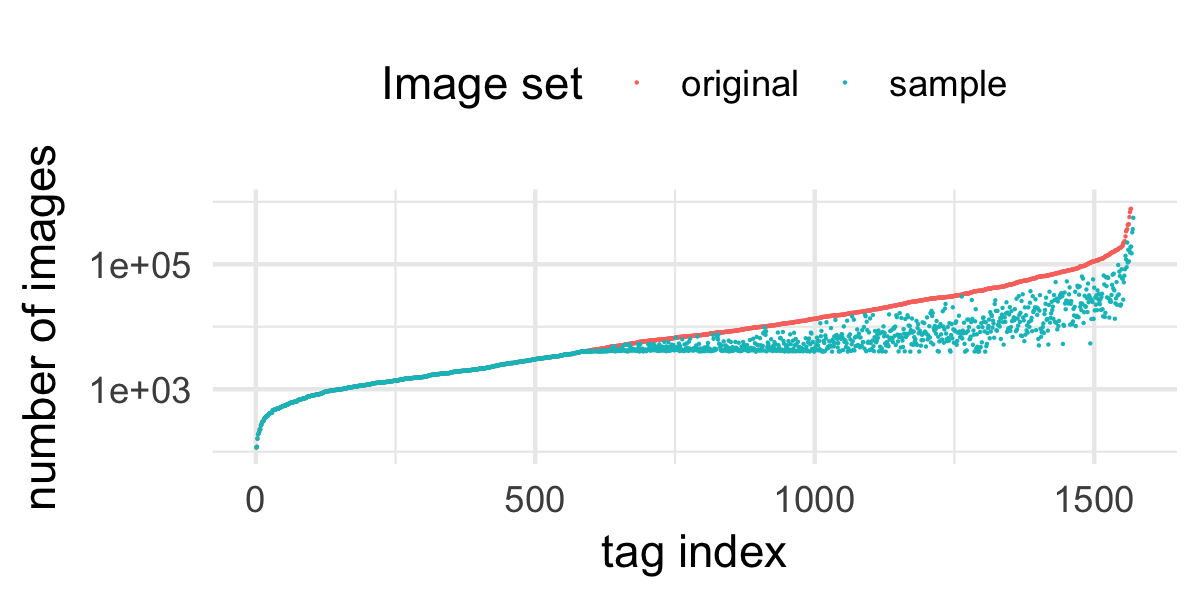}
\vspace{-5pt}
\caption{Sampling 1.0 from 4.8 million images. The tags were sorted according to increasing frequency in the pre-sample set (red). The two histograms start diverging at the minimum quota $Q=4000$ images per tag. Ideally, the rest of the (green) histogram should be flat, however this is not achieved as an image can have multiple tags.\vspace{-20pt}}
\label{fig:tag_sampling}
\end{figure}

In the first stage we selected images with an appropriate Creative Commons (CC) license that allows editing and redistribution, and chose those with  available machine tags (from YFCC100m) and a resolution between $960\times540$ and $6000\times6000$. From this set of 4,807,816 images, we proposed a new tag-based sampling procedure that was used to select one million images such that their machine tag distribution covers  the larger set well, see Fig. \ref{fig:tag_sampling}.

In the second stage, all images in the set of one million, that were larger than  $1024\times768$ were downloaded and rescaled to $1024\times768$ pixels, while cropping was applied to maintain the pixel aspect ratio. In order to keep faces in the frame, as well as salient parts of the image we designed our own cropping method. It relied on the Viola-Jones face detector and the saliency method of Hou et al. \cite{hou_image_2012}. 13,000 images were then sampled while enforcing a uniform distribution across eight image indicators. Duplicates were removed, using a sampling strategy that accounts for content and indicators. This collection was manually filtered for inappropriate content resulting in our KIQ-10 dataset of 10,073 images.

\subsection{Initial tag-based content sampling}

Downloading 4.8 million images consumes much bandwidth and storage space. Hence, we devise a way to subset 1 million images such that not to reduce their content diversity. We aim at full coverage of content, i.e., having at least one image for each of the 1,570 different machine tags available. To assure a ``uniform’’ coverage, our sample should provide a similar number of images for each tag. This is generally not precisely possible as images have more than one tag (9.2 on average). Therefore, we devised a simple and computationally efficient  sampling heuristic, with the above objectives in mind.
\begin{figure*}[t]
\centering
\begin{minipage}{0.121\linewidth}
\centerline{\includegraphics[width=\textwidth]{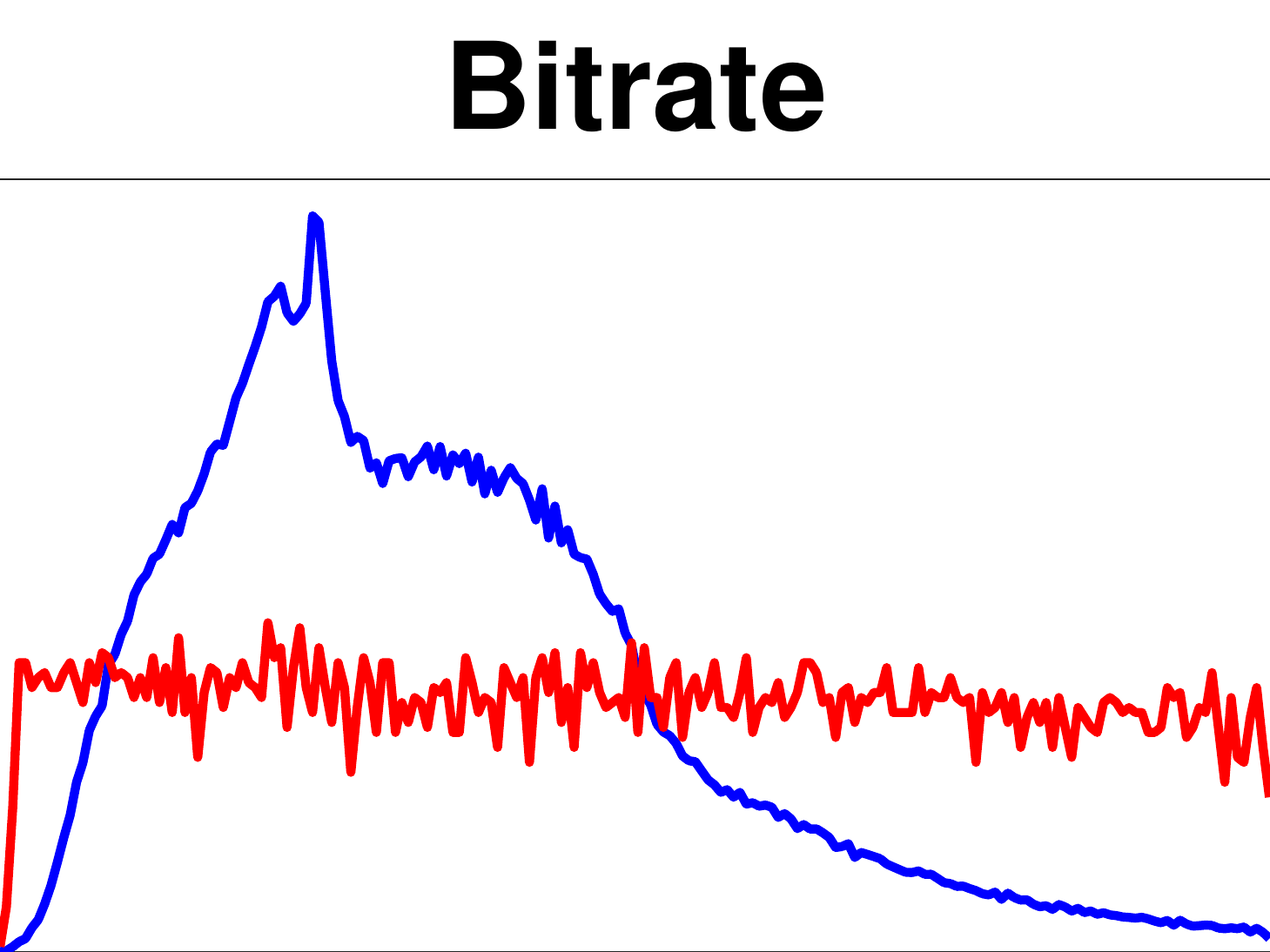}}
\end{minipage}
\begin{minipage}{0.121\linewidth}
\centerline{\includegraphics[width=\textwidth]{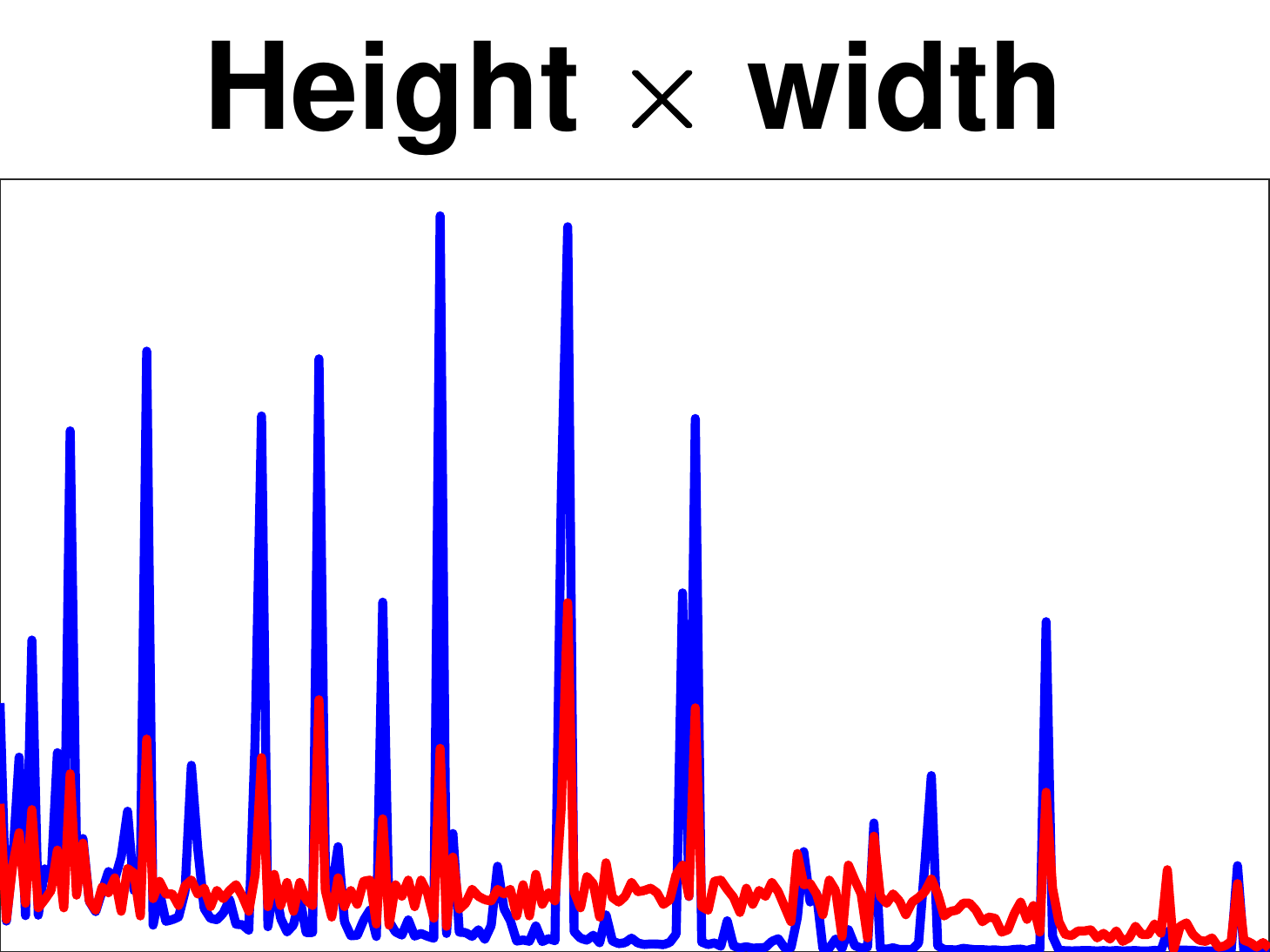}}
\end{minipage}
\begin{minipage}{0.121\linewidth}
\centerline{\includegraphics[width=\textwidth]{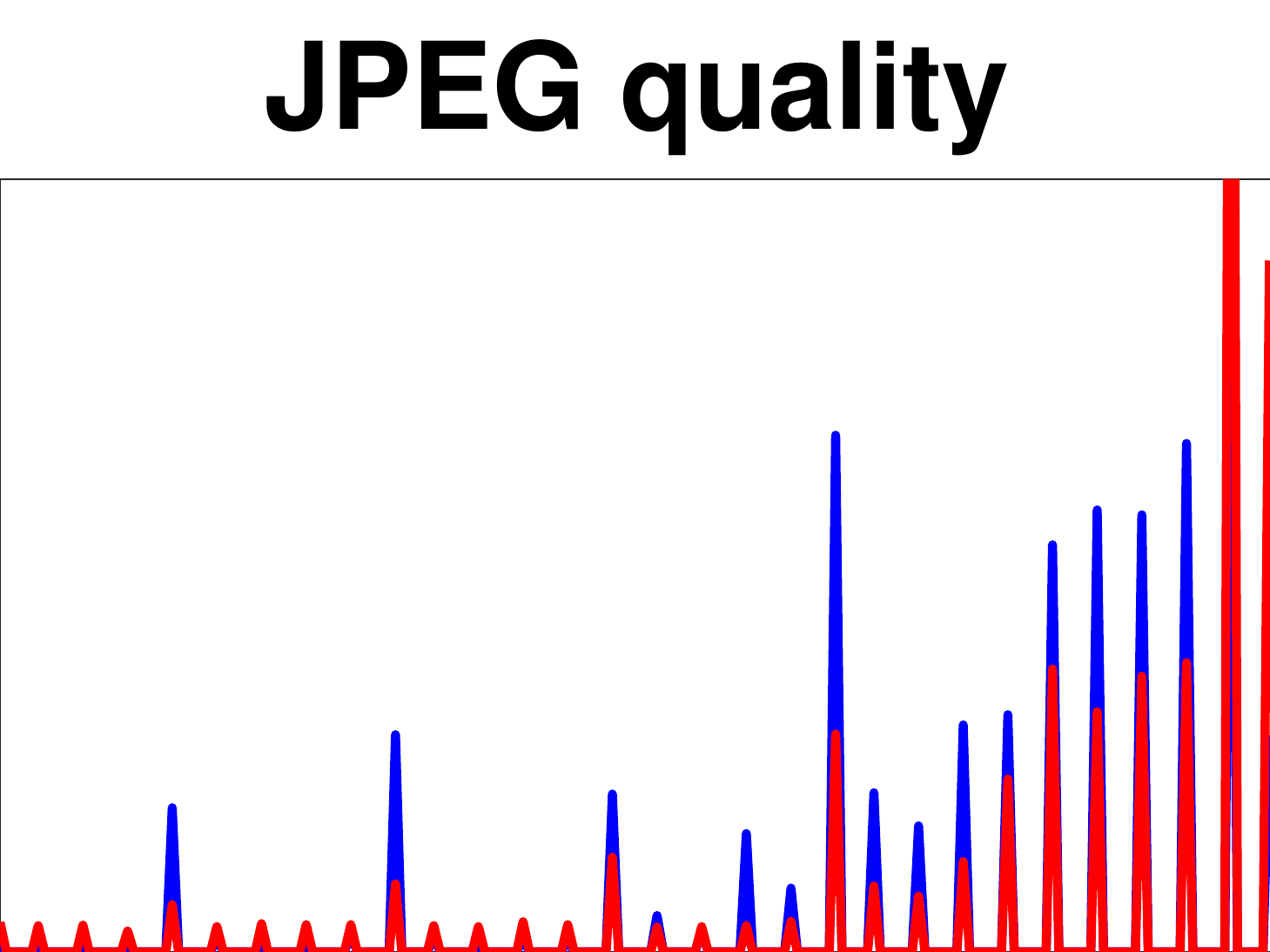}}
\end{minipage}
\begin{minipage}{0.121\linewidth}
\centerline{\includegraphics[width=\textwidth]{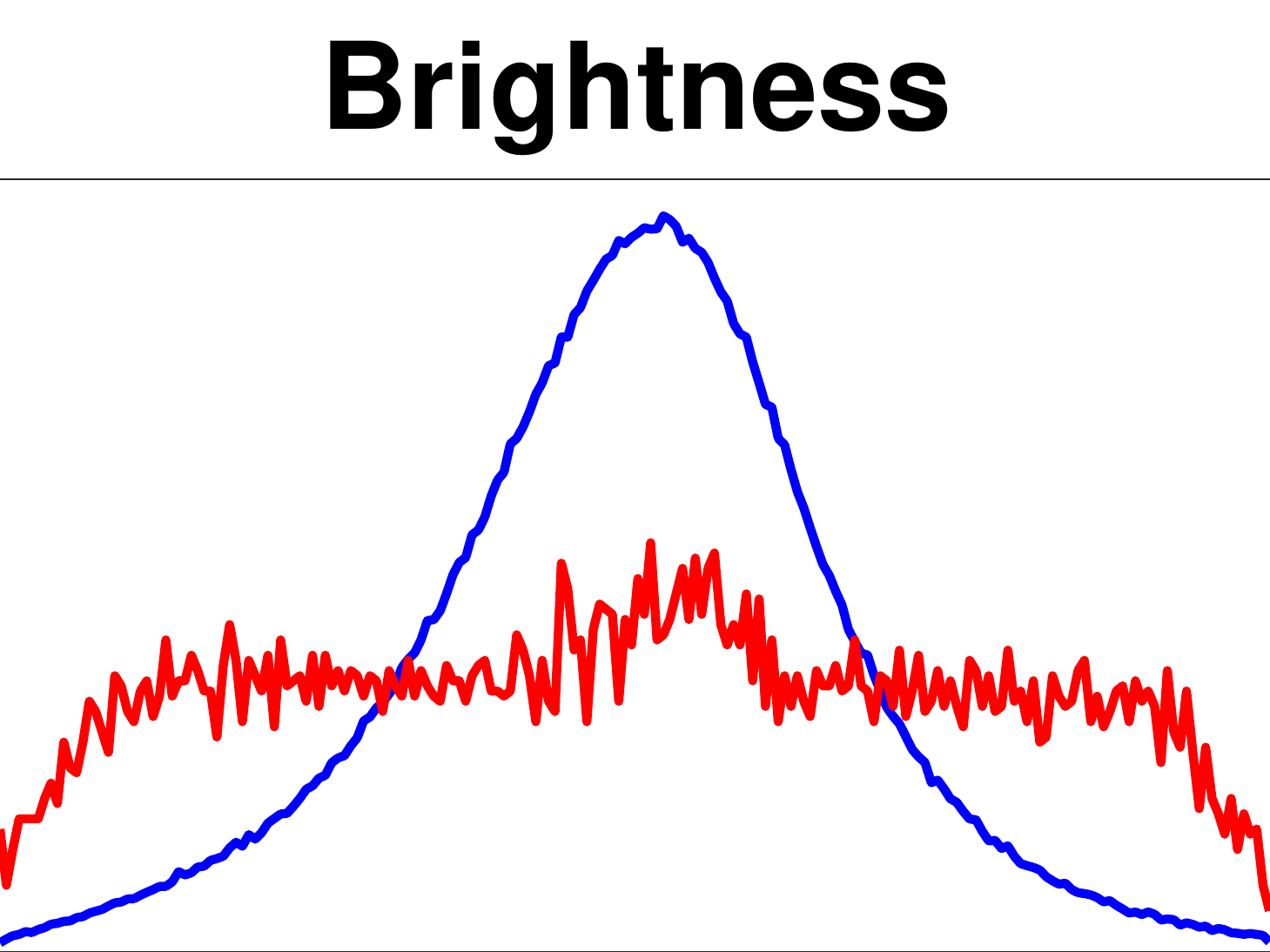}}
\end{minipage}
\begin{minipage}{0.121\linewidth}
\centerline{\includegraphics[width=\textwidth]{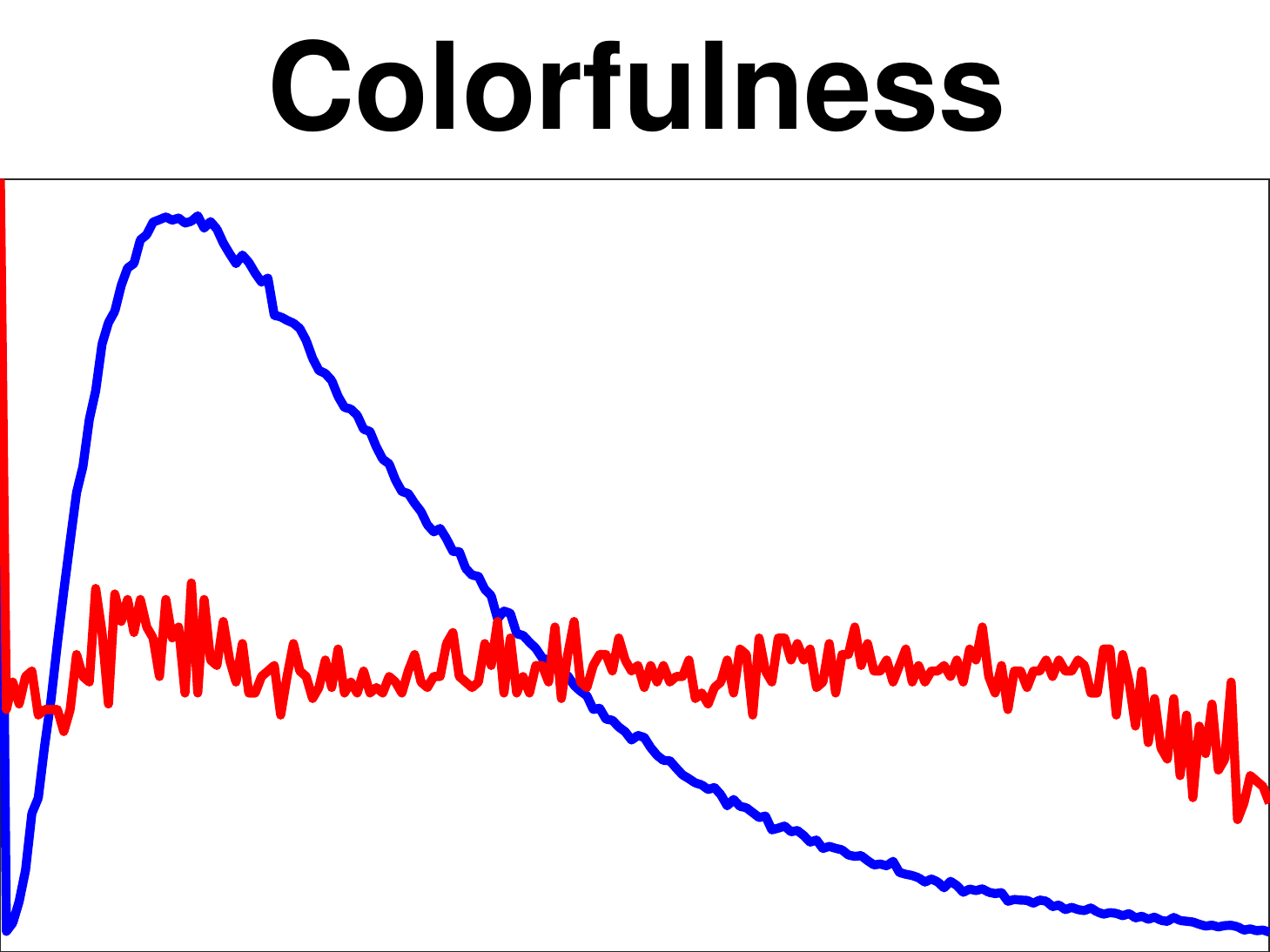}}
\end{minipage}
\begin{minipage}{0.121\linewidth}
\centerline{\includegraphics[width=\textwidth]{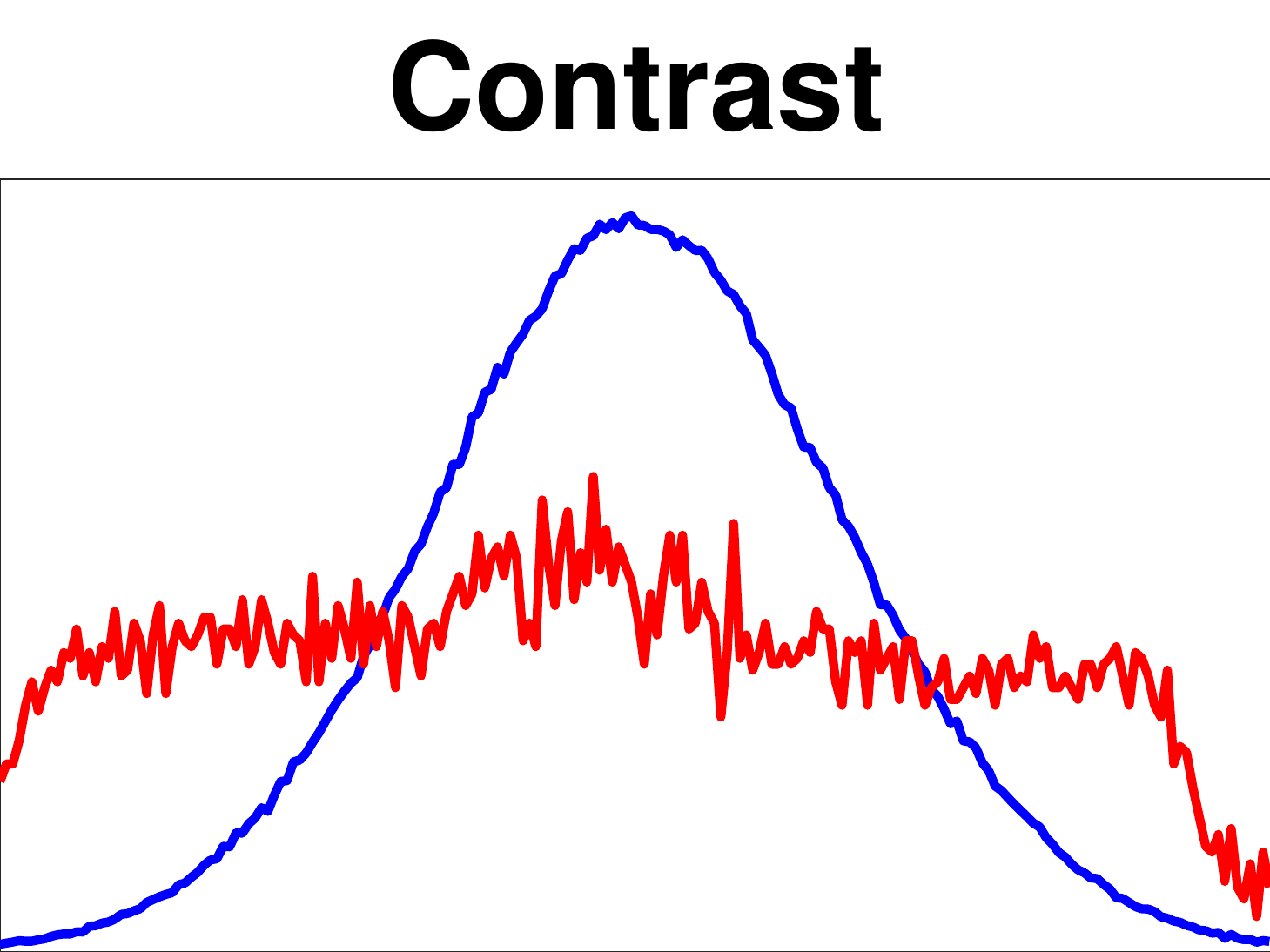}}
\end{minipage}
\begin{minipage}{0.121\linewidth}
\centerline{\includegraphics[width=\textwidth]{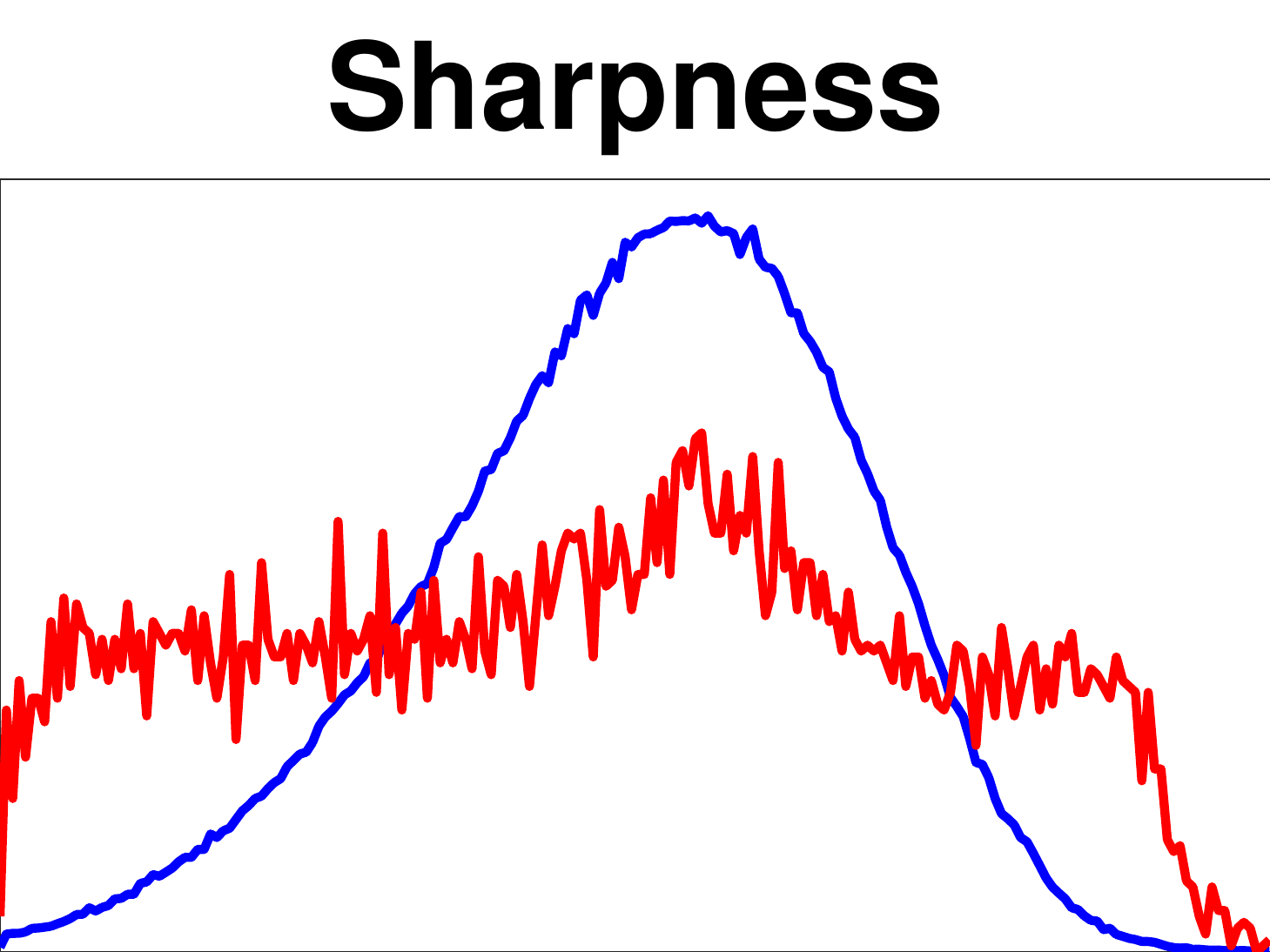}}
\end{minipage}
\begin{minipage}{0.121\linewidth}
\centerline{\includegraphics[width=\textwidth]{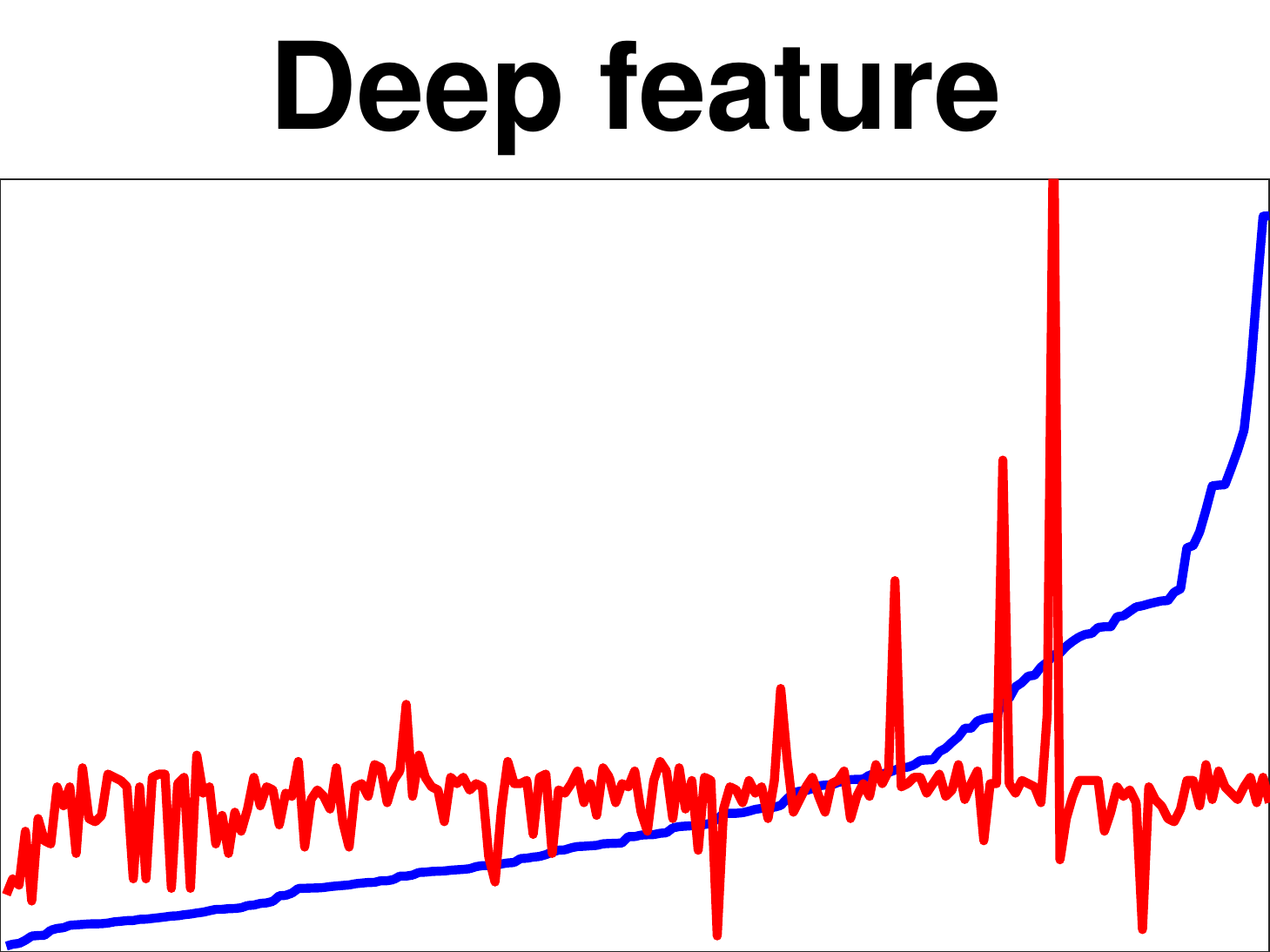}}
\end{minipage}
\caption{Indicator distributions in 866,976 YFCC100m images (blue curves) and sampled 10,073 images (red curves). While the original distributions are far from uniform, the sampling procedure enforces a more uniform distribution on each indicator.}
\label{fig:unisample}
\end{figure*}

Considering the scale of the problem, we propose a computationally efficient method to find an approximate solution. Let $\Phi(t, S_O)$ be the number of images that contain tag $t$ in the set $S_O$ of 4.8 million. We choose a tag quota $Q$ such that all images that contain a tag $t$ with $\Phi(t, S_O) < Q$ are added to the sampled set $S_S$. Let $T(S)$ be the set of tags in a set of images $S$.  For remaining tags $T_R = T(S_O) \backslash T(S_S)$, we include images in $S_S$ such that at least each tag's quota $Q$ is reached. This procedure is as follows. For each tag $t \in T_R$, in  order of increasing counts $\Phi(t, S_O)$, 
we generate an ordered list of candidate images, $O(S_O \backslash S_S, K_t)$, where the list of images is sorted in decreasing order of  $K_t$, the machine confidence in the presence of the tag $t \in T_R$, which is provided by YFCC100m. Then we add the top $Q - \Phi(t, S_S)$ images from $O(S_O \backslash S_S, K_t)$ to $S_S$. To assure that $|S_S| \approx 1,000,000$, one can apply the bisection method to choose the tag quota $Q$. We ran the above algorithm with $Q = 4000$ and stopped adding images to $S_S$ when $|S_S| = 1,000,000 $.

\subsection{Uniform sampling}


To ensure the content diversity and distortion authenticity, we sampled a subset of images while enforcing the uniform distribution across a number of indicators that have impact on image quality and content diversity.

\subsubsection{Indicator selection}
We collected a number of image quality indicators relating to brightness, colorfulness, contrast, noise, sharpness, and No-Reference (NR) IQA measures. Each indicator has at least one implementation. Since we have about 1 million images to evaluate, we dis-considered slow implementations. For the rest of the measures, we conducted preliminary subjective studies and kept four measures that are well correlated with human perception, namely brightness, colorfulness \cite{Hasler:2003}, Root Mean Square (RMS) contrast, and sharpness \cite{Vu:2012}. Besides these, we considered three other indicators: image  bitrate, resolution (height$\times$width), and JPEG compression quality; these are highly correlated with image quality.  
Until this point, we had ensured content diversity by sampling 1 million images based on pre-existing machine tags from YFCC100m. These had been assigned using an existing deep architecture for classification, and represent a few most likely categories per image. To further improve the content description, we rely on the more comprehensive $4096$-dimensional deep features extracted by the pretrained VGG-16 model (FC7) \cite{Simonyan:14c}.

\subsubsection{Sampling strategy}
Each quality indicator identifies an image attribute, measuring its magnitude or presence as a scalar value. Extreme values for an indicator relate to severe distortion, either due to the absence or abnormal emphasis on that particular aspect. If we were to randomly sample our image database, it is unlikely that images having ``abnormal" attribute values would be selected. Therefore, we employed a sampling strategy which generates more images with a wide range of indicator values, and thus more distortions and content types.

Nonetheless, the absolute extremes of the indicator ranges are distorted to an excessive degree, not being informative, e.g., overly dark or bright, overly colorful, etc. Before performing the sampling procedure, we therefore trimmed the extreme ends of each indicator distribution by removing all images with an absolute z-scored indicator value greater than 3. The dataset size shrank from 1 million to 866,976.

For the actual sampling, we applied the method proposed by Vonikakis et al. \cite{vonikakis2016shaping}, enforcing a uniform target distribution for each indicator. The method quantizes each indicator value into $N$ bins. The sampling procedure jointly optimizes the shape of the histograms along all indicator dimensions, using Mixed Integer Linear Programming (MILP).

We used $N=200$ bins for all seven scalar indicators. Since the deep features are 4096-dimensional vectors, we applied a bag-of-words model to quantize them. That is, we ran $k$-means to compute 200 centroids, mapping each deep feature to the nearest cluster. We ran the sampling procedure generating 13,000 images, with uniformly sampled indicators. The set is larger than the target of 10,000 to allow for removing duplicates and other post-filtering. 

\subsection{Removal of duplicates and inappropriate content}

The uniform sampling as described ensures the diversity of the image database at a broad scale. However, due to the binning procedure, identical copies or near-duplicate images can be sampled together, e.g.,  photos of a scene taken from slightly different points of view.

We devise a way to remove near-duplicates. First, the values of each indicator were remapped to the interval $[0,1]$. We computed all pair-wise euclidean distances $D(i,j)$ between images $i,j$ from the source dataset in the 8-dimensional indicator plus content space. The distance in the content space is 0 if two images are part of the same cluster, and 1 otherwise. Duplicate and near-duplicate images $i,j$ are expected to correspond to small distances $D(i,j)$. Thus, by iteratively removing a member of the closest pair, we can effectively remove near-duplicates. We removed 2,000 images in this way. 

To ensure the quality of our database we manually removed images showing too little content, namely text screen shots, text scans, heavily under-exposed, or inappropriate images showing mature content. At the end 10,073 images remained which make up our KonIQ-10k database, see Fig.~\ref{fig:unisample}.

\begin{figure*}[t]
\vspace{-15pt}
\centering
\begin{minipage}{0.2\linewidth}
\centerline{\includegraphics[width=\textwidth,height=70pt]{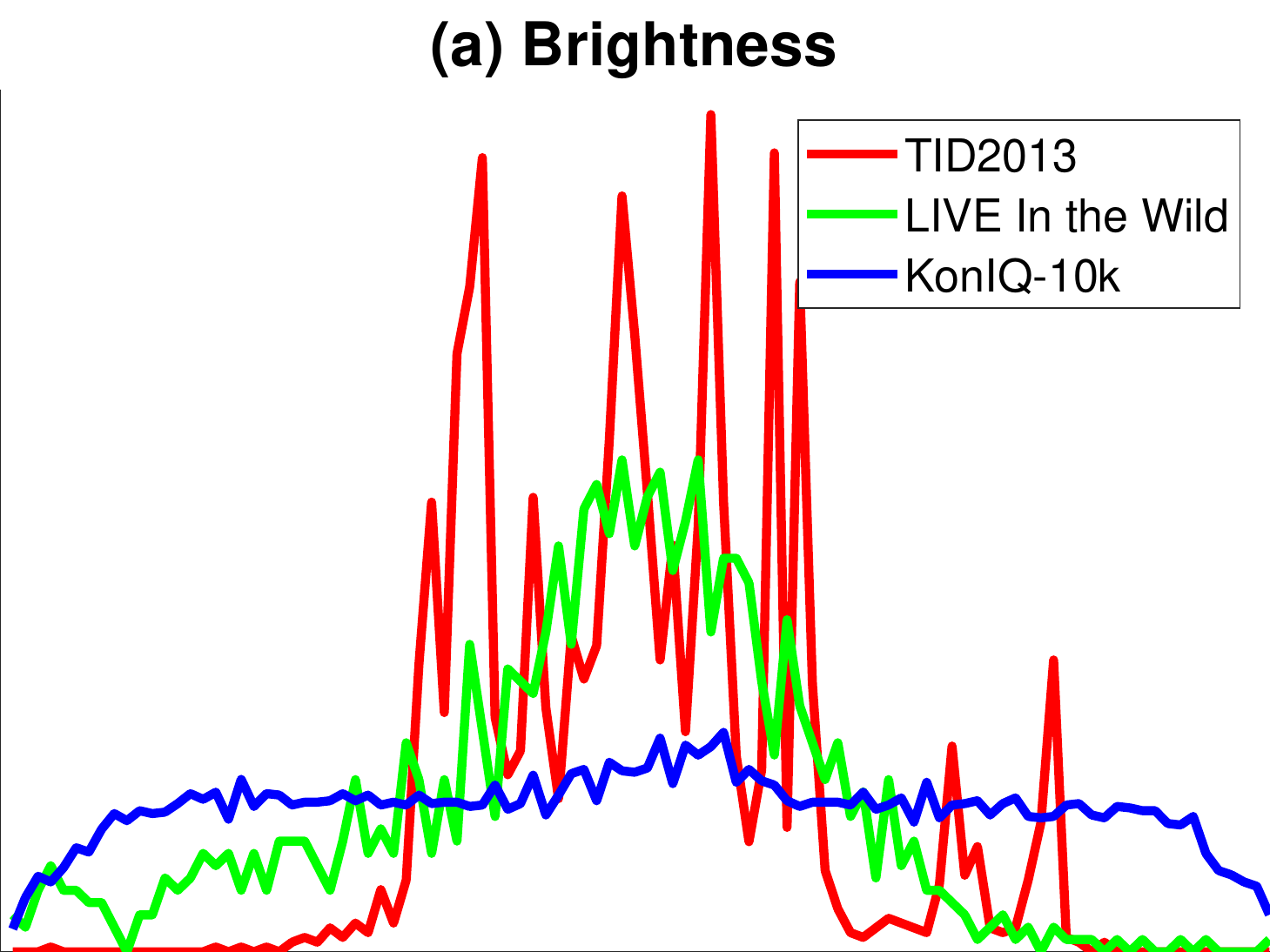}}
\end{minipage}
\hspace{8pt}
\begin{minipage}{0.2\linewidth}
\centerline{\includegraphics[width=\textwidth,height=70pt]{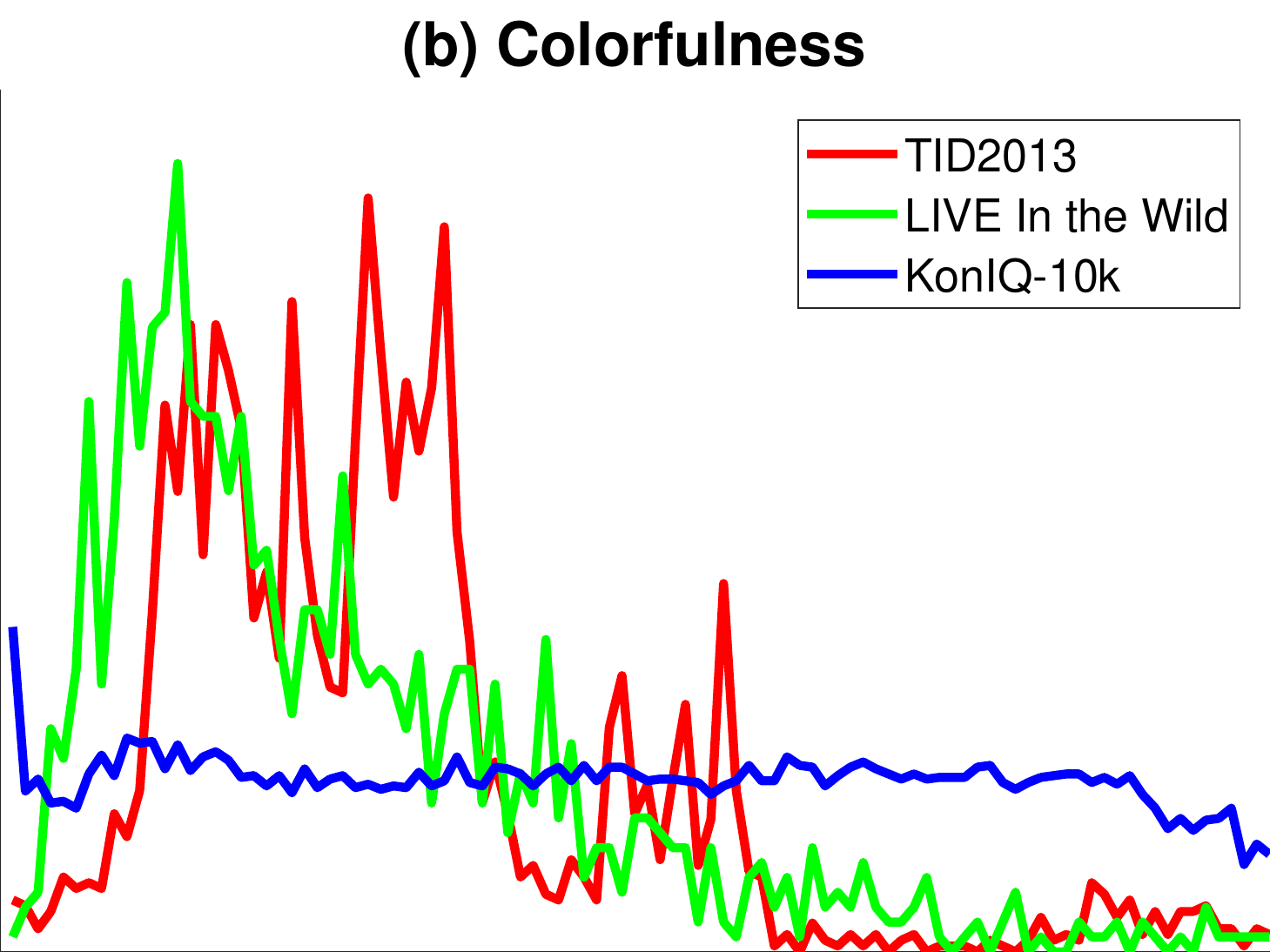}}
\end{minipage}
\hspace{8pt}
\begin{minipage}{0.2\linewidth}
\centerline{\includegraphics[width=\textwidth,height=70pt]{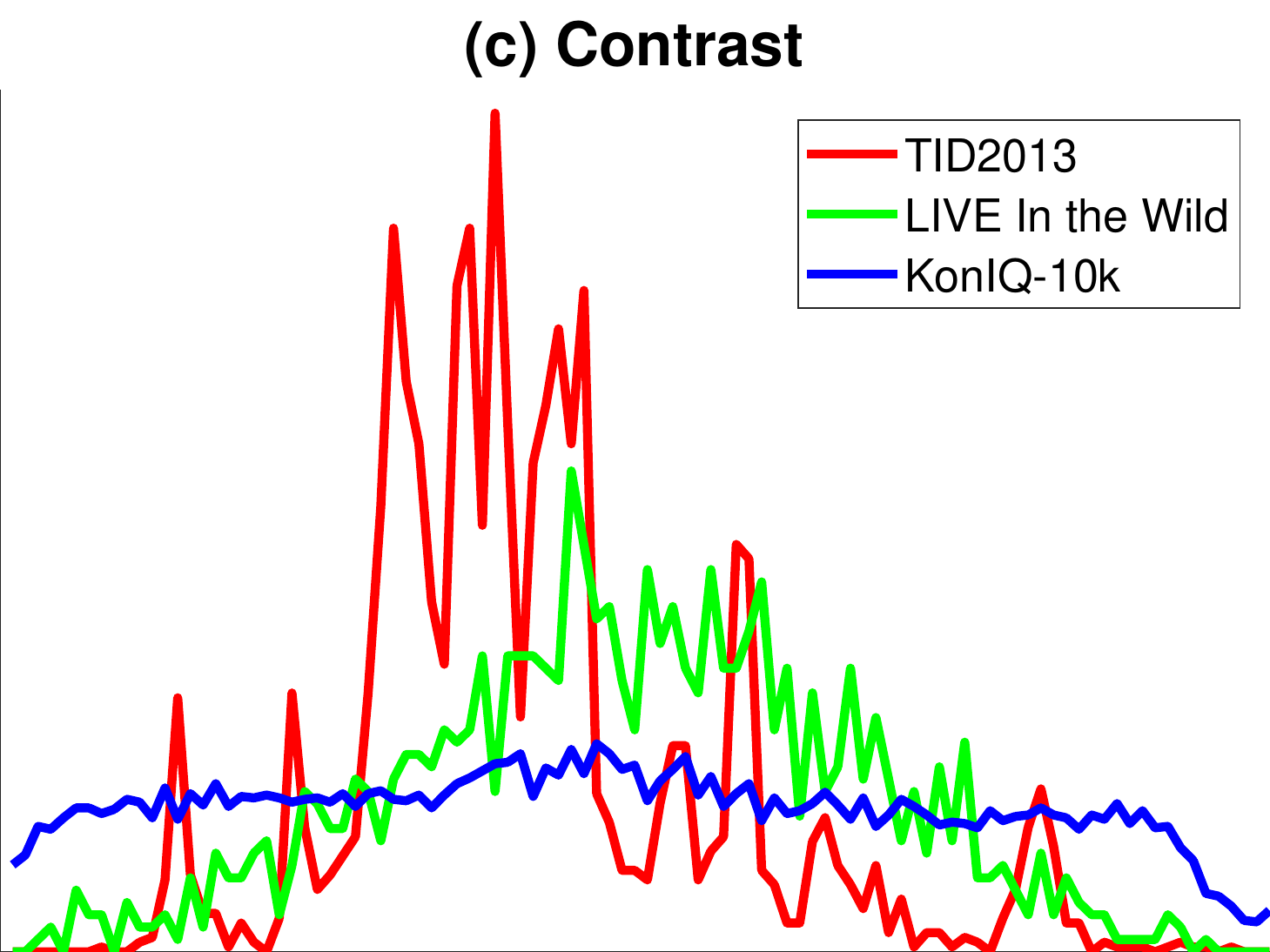}}
\end{minipage}
\hspace{8pt}
\begin{minipage}{0.2\linewidth}
\centerline{\includegraphics[width=\textwidth,height=70pt]{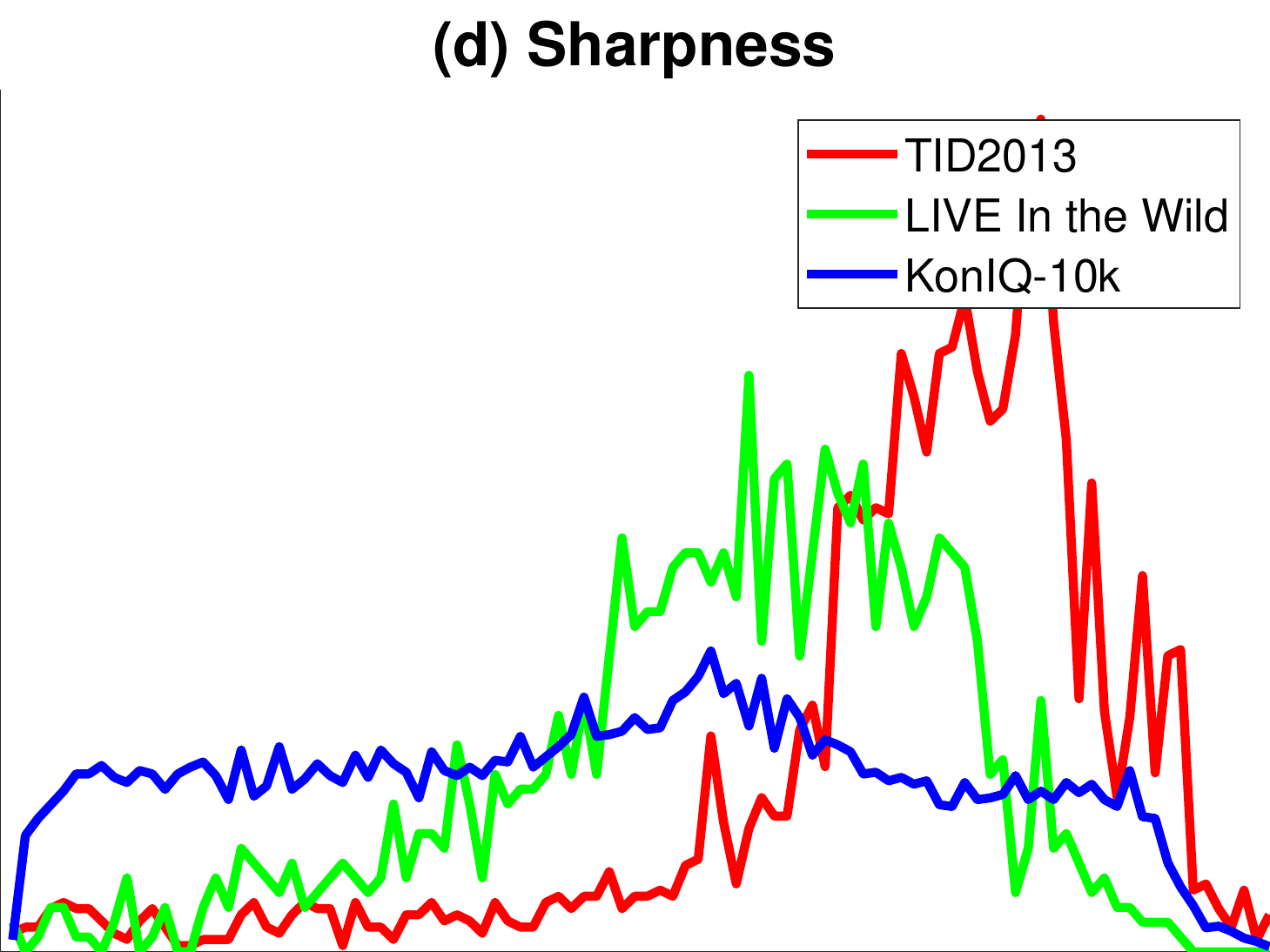}}
\end{minipage}\\
\vspace{8pt}
\begin{minipage}{0.2\linewidth}
\centerline{\includegraphics[width=\textwidth,height=70pt]{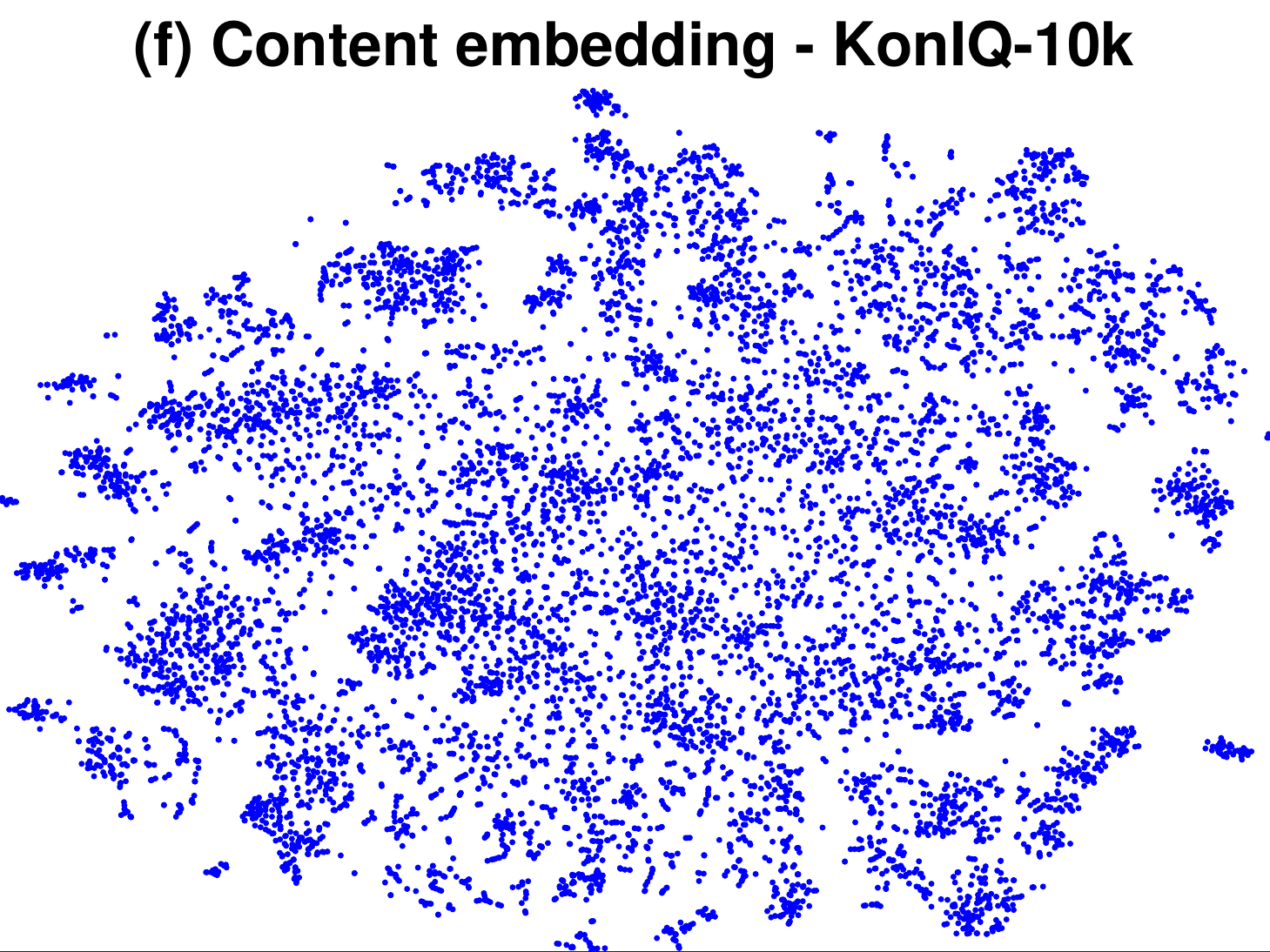}}
\end{minipage}
\hspace{8pt}
\begin{minipage}{0.2\linewidth}
\centerline{\includegraphics[width=\textwidth,height=70pt]{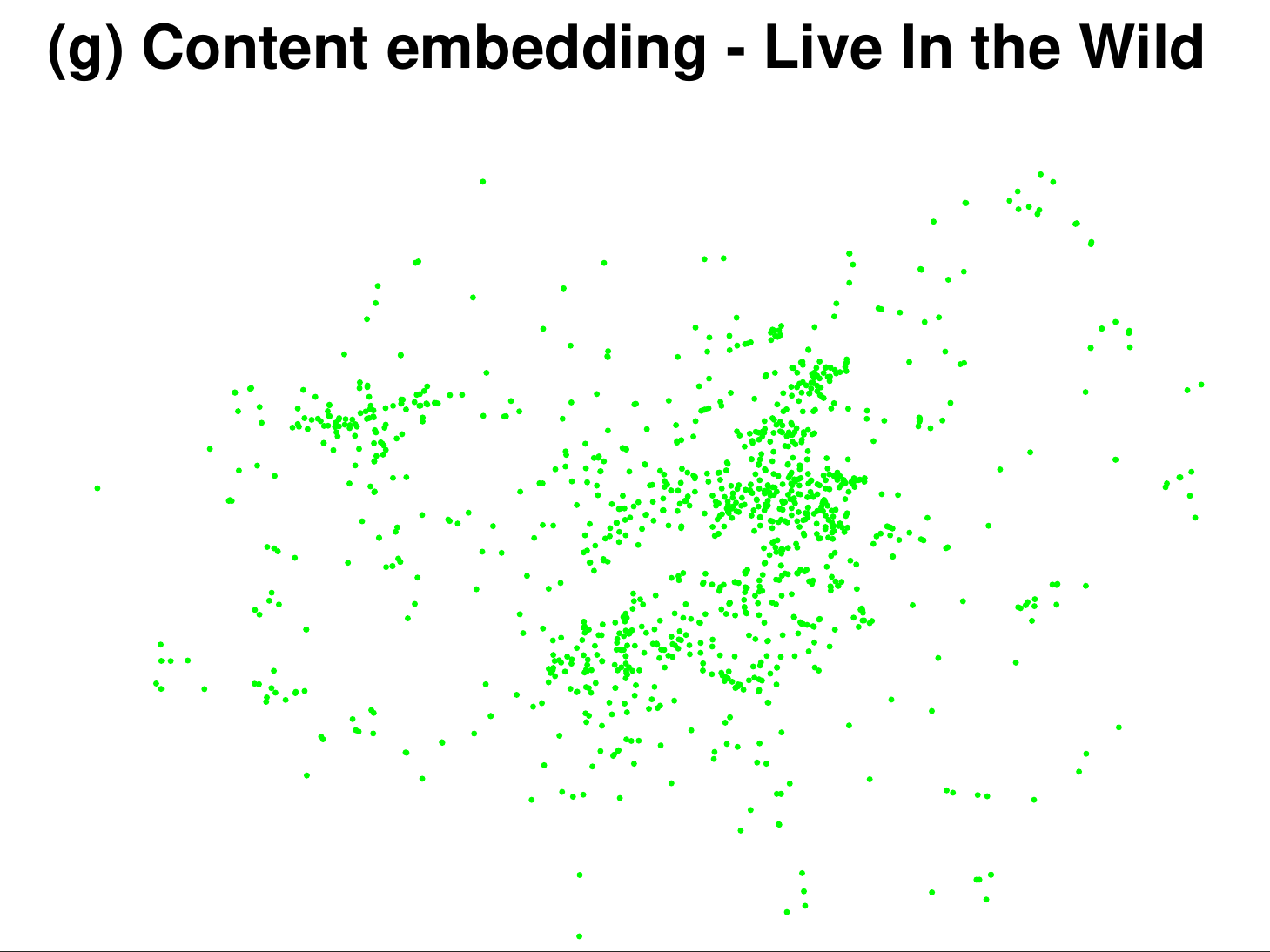}}
\end{minipage}
\hspace{8pt}
\begin{minipage}{0.2\linewidth}
\centerline{\includegraphics[width=\textwidth,height=70pt]{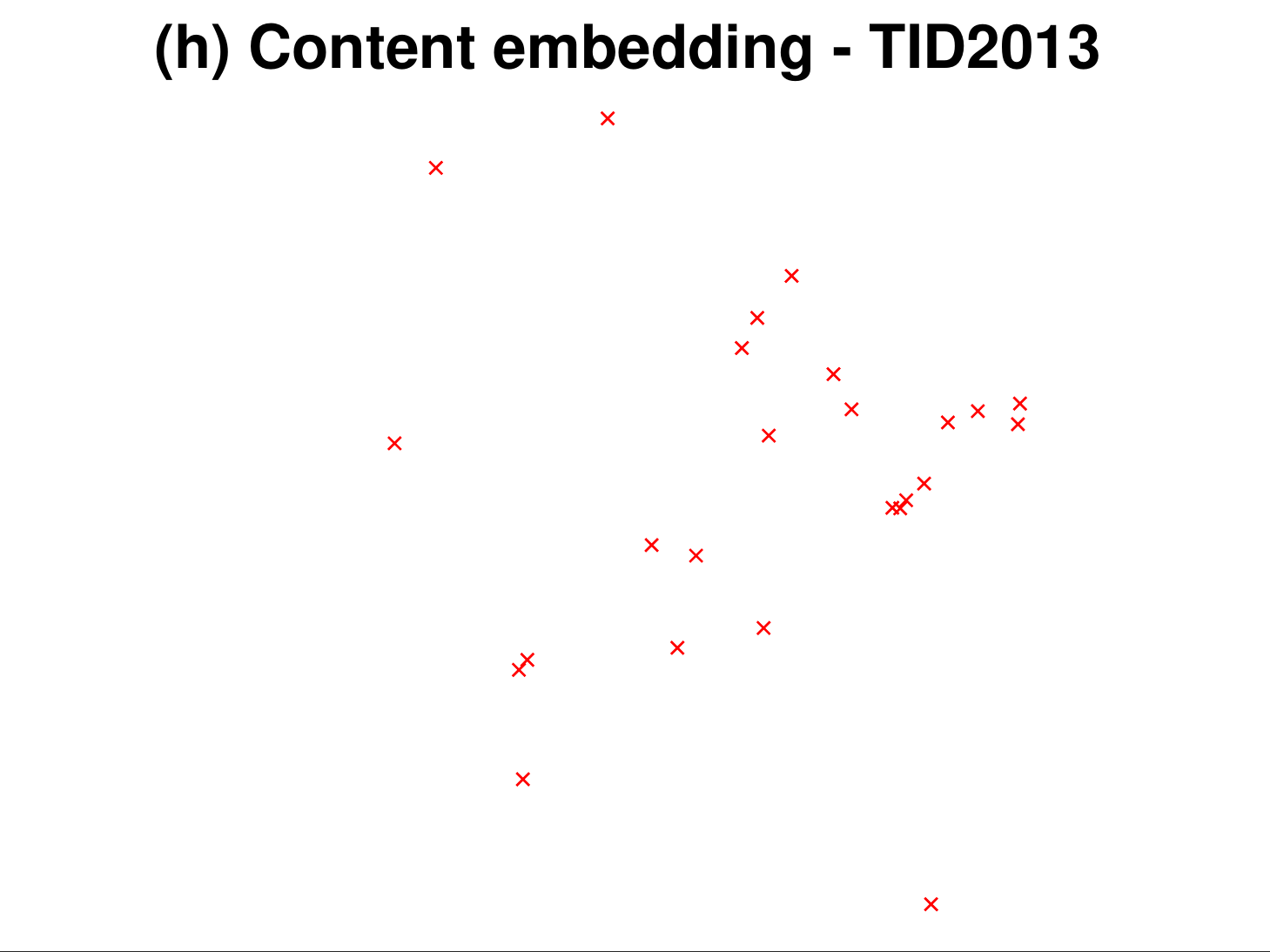}}
\end{minipage}
\hspace{8pt}
\begin{minipage}{0.2\linewidth}
\centerline{\includegraphics[width=\textwidth,height=70pt]{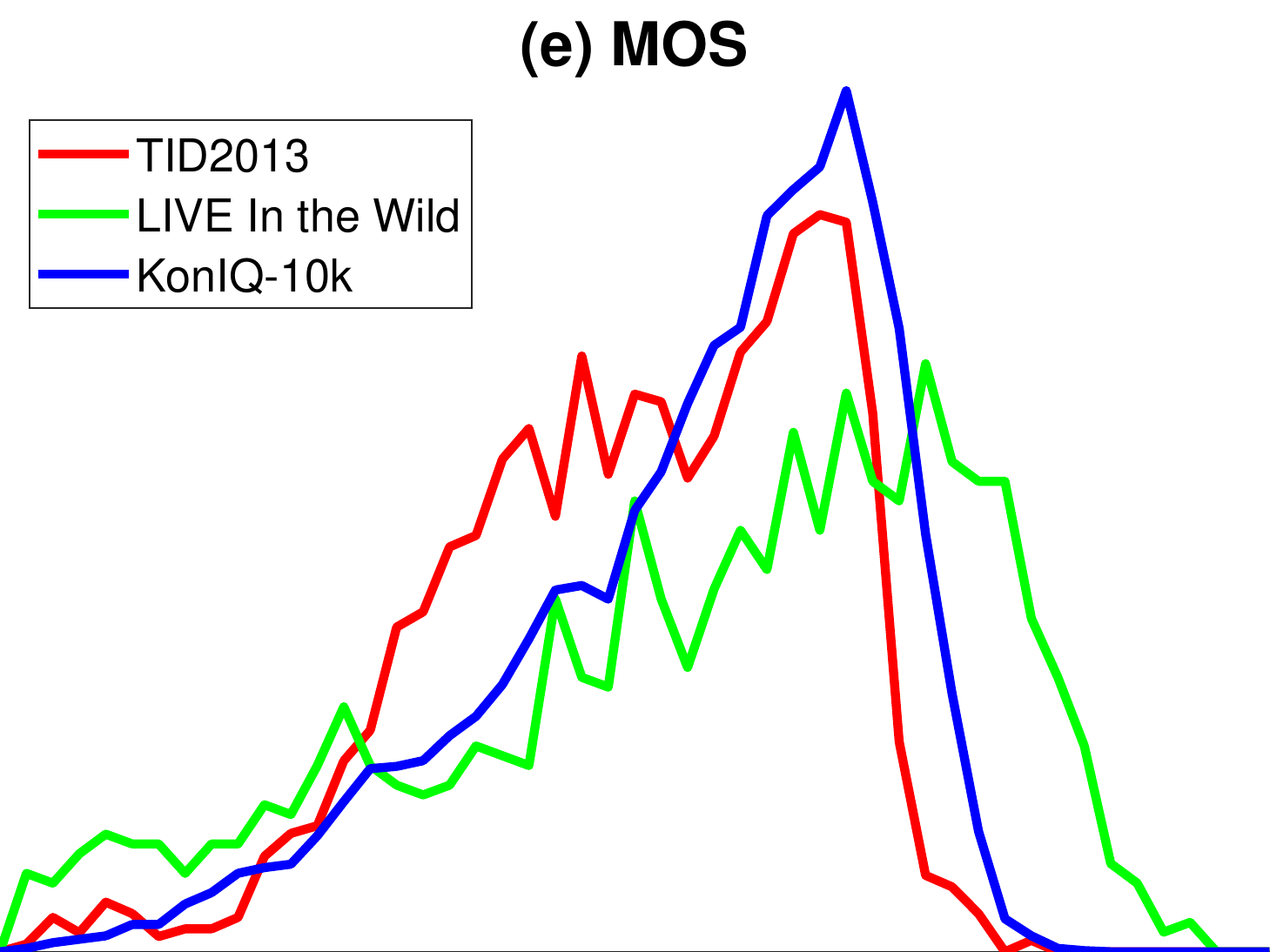}}
\end{minipage}
\caption{Diversity comparison between TID2013, Live In the Wild, and KonIQ-10k. (a) - (d) distribution comparison in brightness, colorfulness, contrast, and sharpness, respectively. (f) - (h) deep feature embedding in 2D via t-SNE. (e) MOS distribution.}
\label{fig:diversitycomp}
\vspace{-10pt}
\end{figure*}

\section{Subjective Image Quality Assessment} 


In order to assess the visual quality of the 10,073 selected images we performed a large scale crowdsourcing experiment on CrowdFlower.com. The experiment first presented workers with a set of instructions, including the definition of technical image quality, considerations when giving ratings, examples of often encountered distortion types, and images with different ratings. 
The subjects were instructed to consider the following types of degradations: noise, JPEG artifacts, aliasing, lens and motion blur, over-sharpening, wrong exposure, color fringing, and over-saturation.
We used a standard 5-point Absolute Category Rating (ACR) scale, i.e., bad (1), poor, fair, good, and excellent (5). Before starting the actual experiment, workers would take a quiz, all questions of which had labeled answers (known as test questions). Only those with an accuracy surpassing 70\% were eligible to continue. Hidden test questions were presented throughout the rest of the experiment, to encourage contributors to always pay full attention. 

The opinions of domain experts are generally more reliable, and thus provide a good source of information for setting test questions. We involved 11 freelance photographers, who had on average more than 3 years of professional experience. We asked them to rate the quality of 240 images: 29 were pristine high quality images, carefully selected beforehand, 21 were artificially degraded using 12 types of distortions and the remaining 190 images were randomly selected from Flickr (not part of our 10k dataset). The distortions included blur, artifacts, contrast, and color degradation. Based on this set of images and the mean opinion score from the freelancers, we generated test questions for our crowdsourcing experiment. The correct answers were based on the rounded values of the freelancers' MOS $\pm$ one standard deviation. All images had at most three valid answer choices.

\section{Results and analysis}

\subsection{Diversity analysis}


We selected LIVE In the Wild and TID2013 to compare their diversity with KonIQ-10k in some aspects. Here LIVE In the Wild and TID2013 are the most representative authentic distorted and artificial distorted databases, respectively. 
Their distributions in brightness, colorfulness, contrast, and sharpness are depicted in Fig.~\ref{fig:diversitycomp}(a)-(d), respectively. Obviously, KonIQ-10k features more diversity in each of those indicators. To compare the content diversity, we embedded the 4,096-dimensional VGG-16 deep features from the databases into a 2D subspace by t-SNE \cite{van2008visualizing}. The visualization is shown in Fig.~\ref{fig:diversitycomp}(f)-(h). Clearly, since LIVE In the Wild images were captured by a few photographers, their content only covered a small region of KonIQ-10k, not to mention TID 2013, generated from only 25 reference images. Their MOS distributions are illustrated in Fig.~\ref{fig:diversitycomp}(e) after rescaling to 1--100 range.



\subsection{Crowdsourcing experiment}

The experiment took more than two weeks to complete. Of 2,302 crowd workers taking the quiz, 1,749 passed it (76\%). For those who passed the quiz and started work, 6\% (101 contributors) failed to meet the 70\% pass rate  on test questions during work. This indicates the quality of our test questions, which were effectively filtering unqualified workers. As a result, to annotate the entire database of 10,073 images, with at least 120 scores each, more than 1.2 million trusted judgments were submitted (over 70\% accuracy). 

In \cite{QoMEXReliability}, the authors have shown that screening users based on image quality test questions improves the intra-class correlation coefficient (ICC), leading to an increased reliability. They have found an improvement from an ICC of 0.37 before screening to 0.5 when users are screened on 70\% accuracy on quality based test questions. The approach in our paper has a similar effect, leading to an ICC of 0.46 on the entire database.

\subsection{Reliability of the crowd}

In order to better study the reliability of the crowd data, we screened workers for unwanted behavior. First, we removed those that had a low agreement with the global mean opinion scores (MOS). Workers that had a PLCC of their votes and the crowd MOS lower than $0.5$ (68 users) were removed.

Second, we detected line-clickers, workers that answered the same too often. We computed the scores' counts of each worker, for all five answer choices. We then took the ratio between the maximum count, and the sum of the four lower counts. Workers with a ratio larger than $2.0$ were removed (121 workers). The MOS for all images was recomputed, after mapping the individual worker scores to $[1,100]$.

\begin{figure}[!htb]
\vspace{-5pt}
\centering
\begin{minipage}{.47\linewidth}
\centerline{(a) average errors}
\centerline{\includegraphics[width=\textwidth]{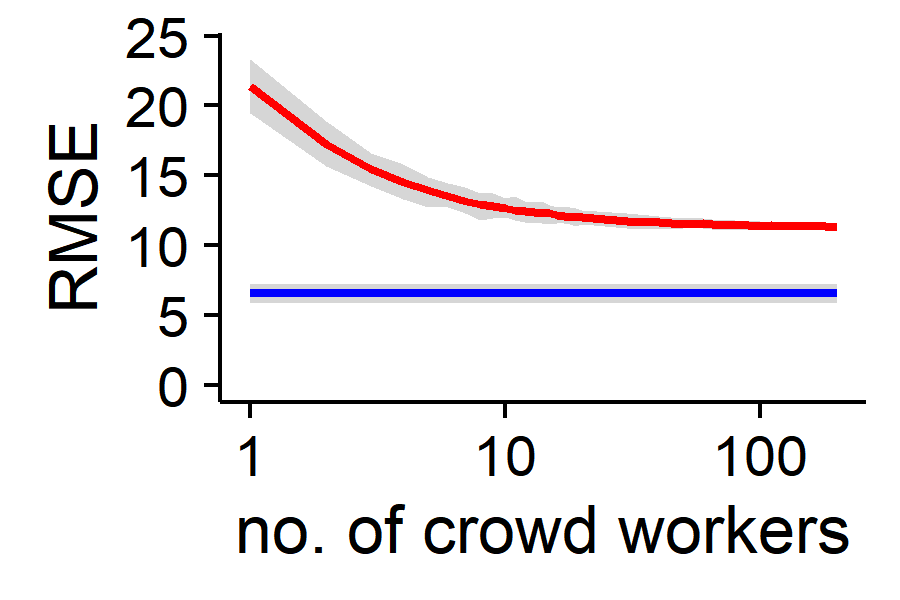}}
\end{minipage}%
\begin{minipage}{.47\linewidth}
\centerline{(b) per image errors}
\centerline{\includegraphics[width=\textwidth]{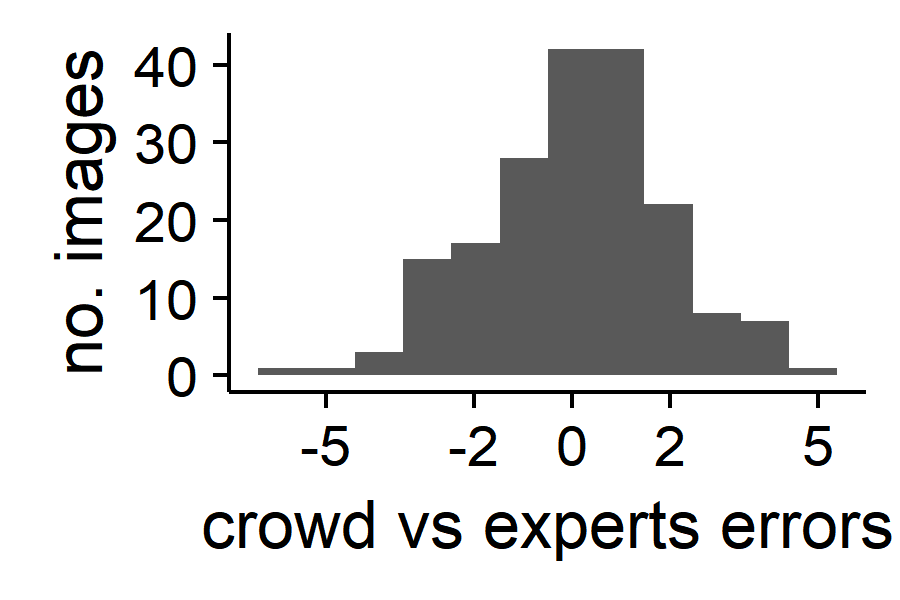}}
\end{minipage}

\caption{(a) Top red line: bootstrapped RMSE of crowd  MOS against 11 experts MOS; Bottom blue line: bootstrapped standard deviation of MOS of 11 experts; gray ribbon is the 95\% CI of the RMSE. (b) Distributions of errors of crowd MOS against experts' MOS, expressed in multiples of the standard deviation of the bootstrapped MOS of 11 experts.}
\label{fig:rmse_curves}
\vspace{-9pt}
\end{figure}

\begin{table}[!htb]
\vspace{-15pt}
\caption{Performance of IQA methods.}
\label{tb:iqacomp}
\centering
\resizebox{0.48\textwidth}{!}{%
\begin{tabular}{l |c c |c c| c c} \hline
& \multicolumn{2}{c|}{KonIQ-10k}& \multicolumn{2}{c|}{LIVE In the Wild}&\multicolumn{2}{c}{TID2013}\\
& SROCC & PLCC&SROCC&PLCC&SROCC&PLCC\\ \hline
BIQI &0.545 &0.619 &0.291&0.388&0.346&0.422\\
BLIINDS-II &0.575&0.583&0.447&0.483&0.529&0.615\\
BRISQUE &0.700& 0.704&0.597&0.630&0.473&0.537\\
DIIVINE &0.585&0.622&0.430&0.468&0.513&0.605\\
SSEQ &0.596&0.615&0.456&0.500&0.510&0.578\\
\hline
\end{tabular}
}
\end{table}

To check the reliability of the crowd MOS, we compare them with those obtained from a group of 11 experts. We have 187 images which have each been rated by 11 experts and at least 592 crowd workers. We compensate for difference in the range of the MOS between the two data sources by fitting a linear model: ${MOS}_{experts} = 1.12\times{MOS}_{crowd} - 10.43$. Relying on this model, the crowd MOS is re-mapped such that relative errors are more indicative of actual performance, and are less affected by changes in scale.

We compare the two data sources: experts and crowd. To do so, we calculate the relative errors between bootstrapped groups of users, by sampling with replacement. We compare the MOS of bootstrapped expert groups of size 11 against the MOS of all 11 experts, and differently sized groups of crowd workers against the MOS of all 11 experts. The crowd sample size varies beyond 120, which is the minimum number of votes we have collected for each of the 10,073 images in our database. In Fig.~\ref{fig:rmse_curves} (a) we show that with respect to errors, crowd workers converge to an agreed MOS quickly (around 30 participants). The crowd opinion is slightly different from that of the expert group, with an RMSE of 11.35 on a 100 point scale. The bootstrapped standard deviation of the experts is 6.63, meaning they also exhibit some inherent disagreement.
 
In Fig. \ref{fig:rmse_curves} (b), we point to the source of the errors by showing their distribution over all 187 images. We note that for a large number of images the errors are within $\pm 2$ standard deviations of the experts' MOS (95\% confidence interval). A crowd MOS value that falls within this interval is likely to have been a result of the votes of 11 experts. We have that 137 of 187 (73\%) images are sufficiently well rated by the crowd so that they can be confused with the ratings of experts. The crowd MOS on the remaining 50 images diverges more from the experts. A preliminary inspection shows that many of the items that have been rated lower by the crowd in comparison to the experts, represent shallow depth of field images (11 of 27). Crowd workers consider the large amount of blur an important degradation, whereas professional photographers understand it as an artistic effect, which doesn't reduce the quality as much. The observed disagreement is at least in part a consequence of diverging domain knowledge between the expert (freelancers) and novice (crowd) groups. Thus, we cannot conclude that the errors, however small, are an indicator for a lower reliability of the crowd.



\vspace{-5pt}
\subsection{NR-IQA evaluation}

We have compared five state-of-the-art NR-IQA  \cite{Moorthy:2010, Saad:2012, Mittal:2012, Moorthy:2011, Liu:2014b} methods on KonIQ-10k, LIVE In the Wild, and TID2013.
We cross-validated a Support Vector Regression model (RBF kernel) on each database using 80\% training / 20\% test set, with 100 repetitions. The average Spearman Rank Order Correlation Coefficient (SROCC) and Pearson Linear Correlation Coefficient (PLCC) are reported in Table~\ref{tb:iqacomp}. 
We observe a wide gap in performance between the two naturally distorted datasets (KonIQ-10k and LIVE In the Wild). An experiment w.r.t.\ size of the database showed that size matters, meaning that larger training sets improve quality predictions which explains the better performance on KonIQ-10k.
\vspace{-3pt}
\section{Summary}

We proposed a new systematic and scalable approach to create an ecologically valid IQA database, KonIQ-10k. To ensure the diversity in content and quality factors, 10,073 images were sampled from around 4.8 million YFCC100m images by enforcing a roughly uniform distribution across seven quality indicators, one content indicator and machine tags. Experimental analysis demonstrated KonIQ-10k is far more diverse than state-of-the-art databases. For each image 120 quality ratings were obtained via crowdsourcing performed by a total of 1,467 crowd workers. We established the quality of the scoring procedure and the reliability of our results with respect to expert ratings. For a more detailed study on this issue see \cite{QoMEXReliability}. Blind IQA is still a challenging task, especially for natural, not artificially distorted images, which calls for IQA databases with natural images like ours. We hope our approach will enable the scientific community to design better and larger databases in the future. Moreover, our dataset KonIQ-10k already has facilitated the design of new blind IQA methods using deep learning \cite{DeepRN,PatchNet}.

\section*{Acknowledgment}
The authors would like to thank the German Research Foundation (DFG) for financial support within project A05 of SFB/Transregio 161. 

\bibliographystyle{IEEEbib}
\bibliography{refs}
\appendix 
\newif\ifsupplementary

\end{document}